# WEM-GAN: Wavelet transform based facial expression manipulation

**Dongya Sun, Yunfei Hu, Xianzhe Zhang, Yingsong Hu**

School of Computer Science and Technology, Huazhong University of Science and Technology, Wuhan, 430074, China

Corresponding author: Yingsong Hu (e-mail: huys@hust.edu.cn)

## Abstract

Facial expression manipulation aims to change human facial expressions without affecting face recognition. In order to transform the facial expressions to target expressions, previous methods relied on expression labels to guide the manipulation process. However, these methods failed to preserve the details of facial features, which causes the weakening or the loss of identity information in the output image. In our work, we propose WEM-GAN, in short for wavelet-based expression manipulation GAN, which puts more efforts on preserving the details of the original image in the editing process. Firstly, we take advantage of the wavelet transform technique and combine it with our generator with a U-net autoencoder backbone, in order to improve the generator's ability to preserve more details of facial features. Secondly, we also implement the high-frequency component discriminator, and use high-frequency domain adversarial loss to further constrain the optimization of our model, providing the generated face image with more abundant details. Additionally, in order to narrow the gap between generated facial expressions and target expressions, we use residual connections between encoder and decoder, while also using relative action units (AUs) several times. Extensive qualitative and quantitative experiments have demonstrated that our model performs better in preserving identity features, editing capability, and image generation quality on the AffectNet dataset. It also shows superior performance in metrics such as Average Content Distance (ACD) and Expression Distance (ED).

**Keywords** Autoencoder, GANs, Facial expression editing, Wavelet transform

# 1 Introduction

Facial expressions are a kind of high-level semantic feature of human faces. It could convey the sensibilities and intention of mankind, which makes it a common but crucial way of communication. In the current era of social media, editing human facial expressions without changing the identity has been widely adopted in many different fields like VR, gaming, filming, and streaming, etc. Facial expression editing technology can be combined with multimodal biometric authentication technology to enhance system security and accuracy [1]. Facial expression editing can simulate various expression changes, providing abundant training data for multimodal biometric systems, thereby improving the system's robustness to expression variations [2]. Additionally, facial expressions can serve as an extra biometric feature to enhance the security of the authentication process [3]. Data augmentation through facial expression editing technology [4] can generate more facial expression samples, which can help train models to better understand and recognize the expression features of different individuals. This augmented dataset can not only improve the model's generalization ability but also reduce overfitting issues caused by insufficient data [5].

With the proposal of Generative Adversarial Networks (GANs) [6], great progress has been made in facial expression editing by using the powerful generation capabilities of GANs in some recent studies. Some researchers take advantage of the GANs to decouple the latent embeddings of the facial expression from the latent embeddings of the human face images and edit the latent embedding to generate the final output [7],[8][9]. These methods do not need any labels, while they could depend on the pretrained generators to yield high-resolution images. Some other studies rely on the Action Units (AUs) labels [10], which are used to train the model to generate the edited expression images [11],[12],[13]. These methods are good for both continuous editing with the AUs labels and keeping the identity information. Although the aforementioned facial expression editing models have demonstrated robustness in facial expression editing tasks, they still have not addressed some challenging issues. The first challenge is fidelity: these models typically generate images with low fidelity, making it challenging to keep the original detail information. The second challenge is the issue of average expressions: the models tend to generate average expressions from the dataset to handle various situations, leading to a lack of personalization in the generated images' expressions.

All the methods mentioned above require the mapping of input images to low dimension latent embeddings through encoder, which are later modified and reconstructed by a decoder to complete facial expression editing. However, based on the Rate-Distortion theory [14], there must be an inevitable loss of information if a real-world image is transformed to a low-dimensional embedding. Also as suggested by the information bottleneck theory [15], the lost information is mainly the specific semantic details of the images since the deep learning model has the momentum to keep the public information from different domains. Thus, the generated images might be low in fidelity, though they still might preserve most of the identity information of each human being so one could re-identify them. But these generated images fail to save some specific details, which makes the fine-grained editing of facial expression hard to achieve.

In order to address the aforementioned issues, our work focuses on processing detail information during editing. Firstly, we chose a U-Net-based generator as our model backbone, and a Detail Information Transmission (DIT) module based on wavelet transform is proposed in the generator of U-Net since it could extract the high frequency component of the input images. The extracted information will be delivered directly to the decoder through skip connections to enhance the generator's ability of learning high frequency detail information. Secondly, we also implement the high-frequency component discriminator to further constrain the network model and encourage the generator network to generate images with more detail information. Additionally, we put a residual block [16] between the encoder and decoder, and relative AUs are introduced multiple times to enhance the network model's ability to capture AUs, making the generated facial expressions more aligned with the target expressions.

In conclusion, the major contributions of this work are as follows.

(1) We combine the wavelet transform technique into our generator to focus on the processing of detail information of images during editing, and improve the ability of the generator to retain more detailed facial features.

(2) We add a high-frequency component discriminator and use high-frequency domain adversarial loss to optimize the generator's performance to collect detail information.

(3) We use the residual network block, with relative AUs introduced several times, to connect the encoder and decoder, so the gap between the generated expressions and the target expressions can be narrowed.

(4) After extensive qualitative and quantitative experiments, we show that our model is able to successfully transform input facial expressions to the target facial expressions while keeping more identity information.

Section 2 introduces related work on GANs, facial expression editing, and the two common network frameworks used for this task. Section 3 describes our proposed method, with Subsection 3.1 presenting the overall training framework, and Subsection 3.2 detailing the network architecture, including the generator and the two discriminators used. Subsection 3.3 introduces our proposed DIT module, and Subsection 3.4 outlines the loss functions employed. Section 4 provides the qualitative and quantitative results of our proposed experimental methods on commonly used public databases. Finally, in Section 5, we summarize the paper and propose directions for future work.

# 2 Related Work

**Generative Adversarial Networks:** GANs，due to their unique training process and powerful generation capability, have received increasingly more attention upon their proposal. However, a lot of attempts have been made to further optimize the adversarial loss function of GANs to make the training process more stable. Wasserstein Generative Adversarial Networks(WGAN) [17] introduces the Wasserstein distance to measure the distance between two distributions, which solves the problems of gradient disappearance and mode collapse in traditional GANs, and WGAN-GP [18] further improves WGAN training procedure by using Gradient Penalty(GP). Meanwhile, Conditional GAN [19] increases the controllability of the generated images by introducing conditional variables in adversarial training. So far, GANs have demonstrated their powerful capabilities in the areas of image generation [20]-[22], image-to-image transformation [23],[24], and super-resolution [25].

**Facial Expression Editing:** Facial expression editing is a very challenging computer vision task and owing to the powerful generative power of generative adversarial networks, GANs-related facial expression editing framework has achieved good results. StarGAN [26] makes it possible to translate multi-domain images using target labels and a single model that can generate facial expressions with discrete labels. ExprGAN [27] linearly combines discrete expression labels and evenly distributed noise to form a controllable new expression label as condition vector, realizing continuous editing of facial expressions. GANimation [11] uses Action Units (AUs) as expression labels and introduces an attention mechanism to focus on specific editing regions, enhancing the editing ability of facial expressions and achieving continuous adjustment. Cascade EF-GAN [28] focuses on processing the more detailed regions of the face (like mouth, eyes, and nose) to keep the identity-related features, and then uses three cascade generator networks to gradually edit the facial expressions, addressing the artifacts and blurs problem that current models have. Ling [12] et al. achieve a finer-grained level of facial expression editing by changing the conditional labels from absolute AUs to relative AUs and introducing a multi-scale fusion module in the generator. Wang [13] et al. transfer the feature maps from the encoder to the decoder in the generator and use the attention mechanism to fuse the feature maps selectively to retain more identity-related information of the input images. Bodur [29] et al. incorporates the 3d dense information with an additional network that guides the training process by the depth consistency loss. However, compared to the AUs-based supervised framework, 3D dense information is hard to obtain and can only be used to transform expressions in discrete domains.

**General Network Frameworks:** Facial expression editing techniques based on GANs can be broadly categorized into two main types: (1) Direct modification of latent representations: This approach maps facial images to a latent space using an encoder, modifies the latent representations in this space, and then reconstructs the images through a decoder, as shown in Fig. 1. (2) Conditional information-guided methods: This method incorporates additional conditional information during the editing process to guide the generation of latent representations, thereby enhancing the accuracy and controllability of the edits, as shown in Fig. 2.

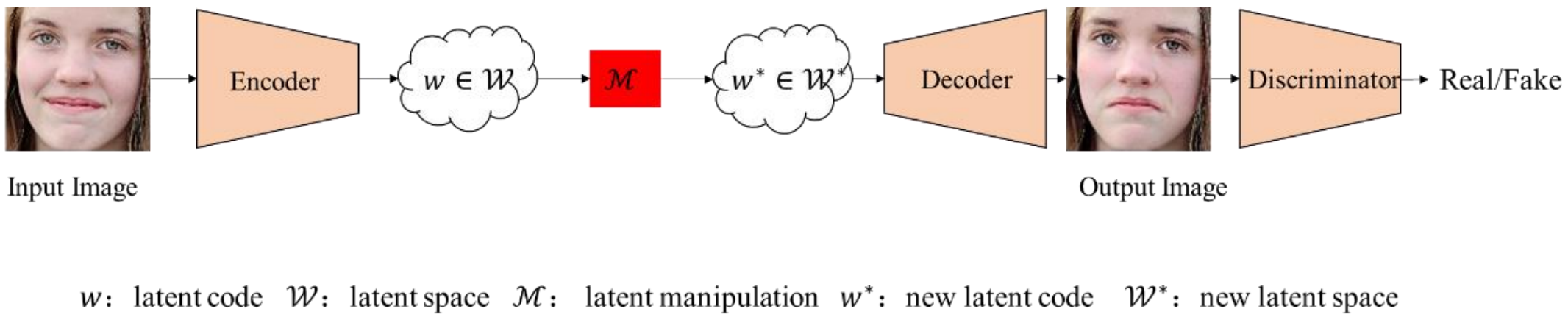

**Fig. 1** Schematic diagram of the network framework for latent representation modification methods

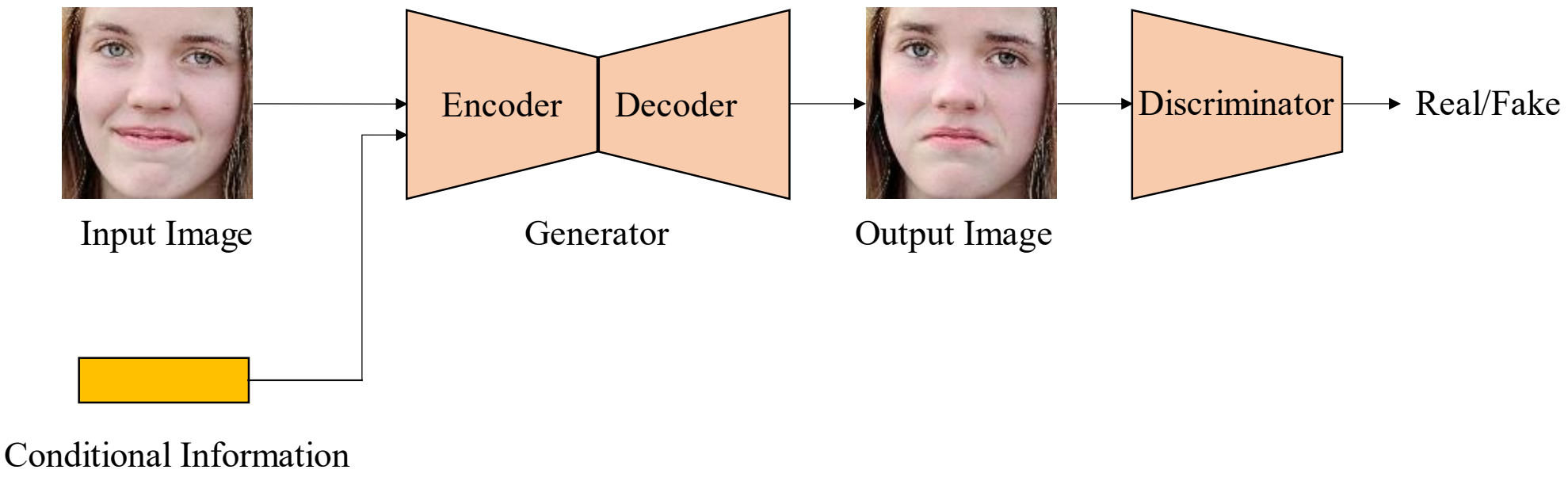

**Fig. 2** Schematic diagram of the network framework for conditional information-guided methods

# 3 Method

In this section, we will illustrate the details of WEM-GAN. We will focus on the loss function and the Detail Information Transmission (DIT) module in our network structure.

## 3.1 Overview

Given an input image $x$ of arbitrary facial expression, we denote its expression labels as $u_x = \{u_x^1, u_x^2 \dots, u_x^n\}$. Given a target image $y$, we denote its expression labels as $u_y = \{u_y^1, u_y^2 \ldots, u_y^n\}$. In these two labels, n is the total number of expression labels, $u_x^i$ is the intensity of the i-th AU, and its value is normalized between 0 and 1. Our goal is to convert the input image into a generated image $x'$ that has the identity information of the input image $x$ and the facial expression of $y$ simultaneously, while keeping as much identity information as possible. We adopt the proposal of Ling [12] et al. to use the relative AUs as the input conditional vector, and the relative AUs represent the change from the original facial expression to the target facial expression, which is denoted as $u_{rel} = u_y - u_x$.

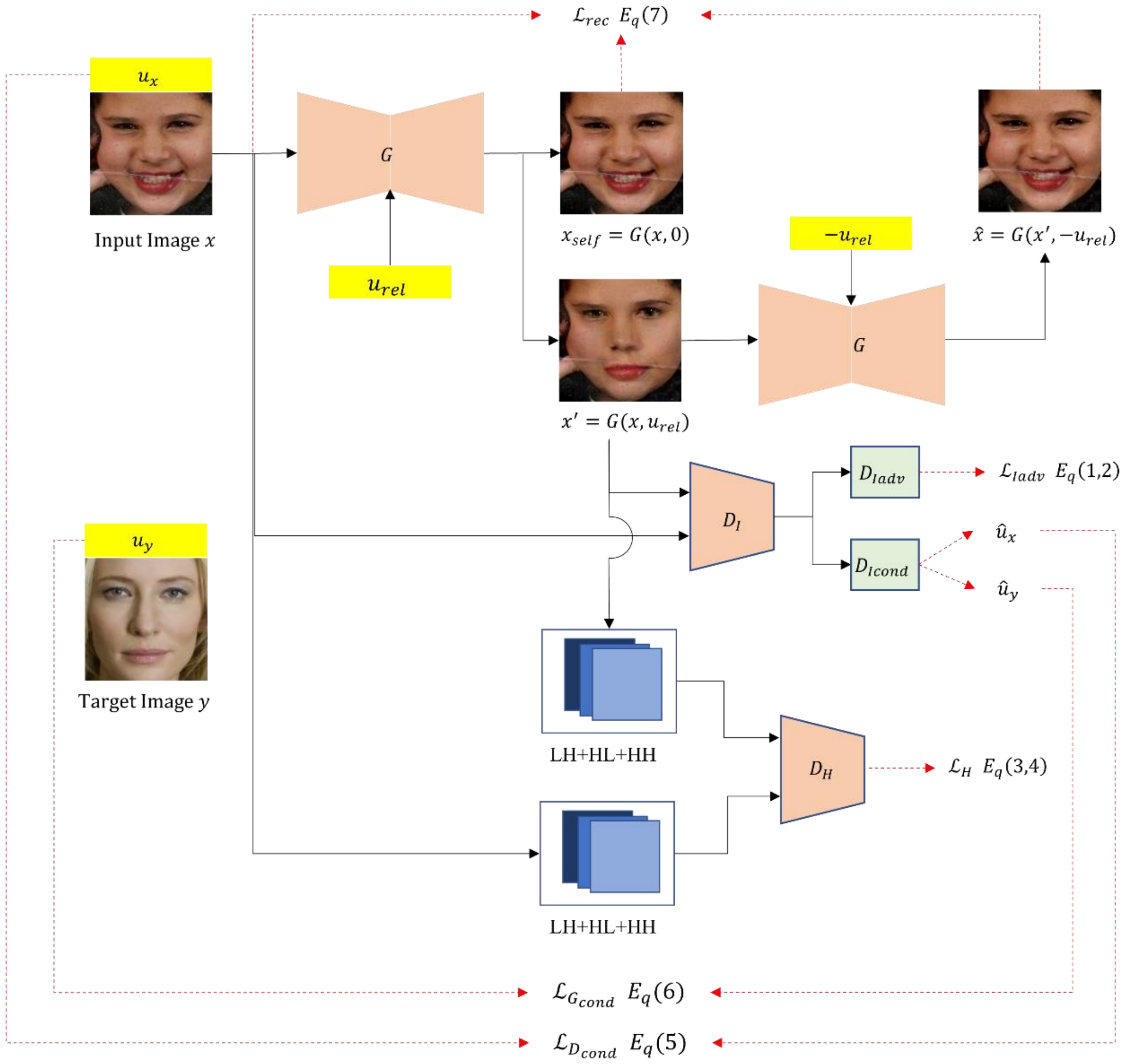

**Fig. 3** Overall framework of our model. Our model consists of a generator $G$ and two discriminators $D_I$ and $D_H$. $G$ edits the input images into images with the target expressions by using the relative AU as a conditional vector and generates the self-reconstructed image and the cycle-reconstructed image during the training process to calculate the identity information loss. The discriminator $D_I$ has two branches ($D_{Iadv}$ and $D_{Icond}$), and $D_{Iadv}$ is used to distinguish the input images from the generated images. $D_{Icond}$ predicts the AUs label of the image, and guarantees that the generated image has the target expression. The discriminator $D_H$ is used to distinguish the detail information of the input images and the generated images. Refer to Section 3.4 for details on $E_q$(1-7) in the diagram.

## 3.2 Network Structure

As shown in Fig. 3, our overall network architecture consists of a generator and two discriminators. The role of the generator is not only to generate fake images with the target expressions, but also to generate self-reconstructed images and cycle-reconstructed images. Then, the information of the images $x$, $x_{self}$ and $\hat{x}$ are used in the cycle consistency loss to ensure the preservation of the identity information of the original image. The role of the discriminator is to ensure the fidelity of the generated images and to supervise whether the generated images have the target expressions and high frequency detail information.

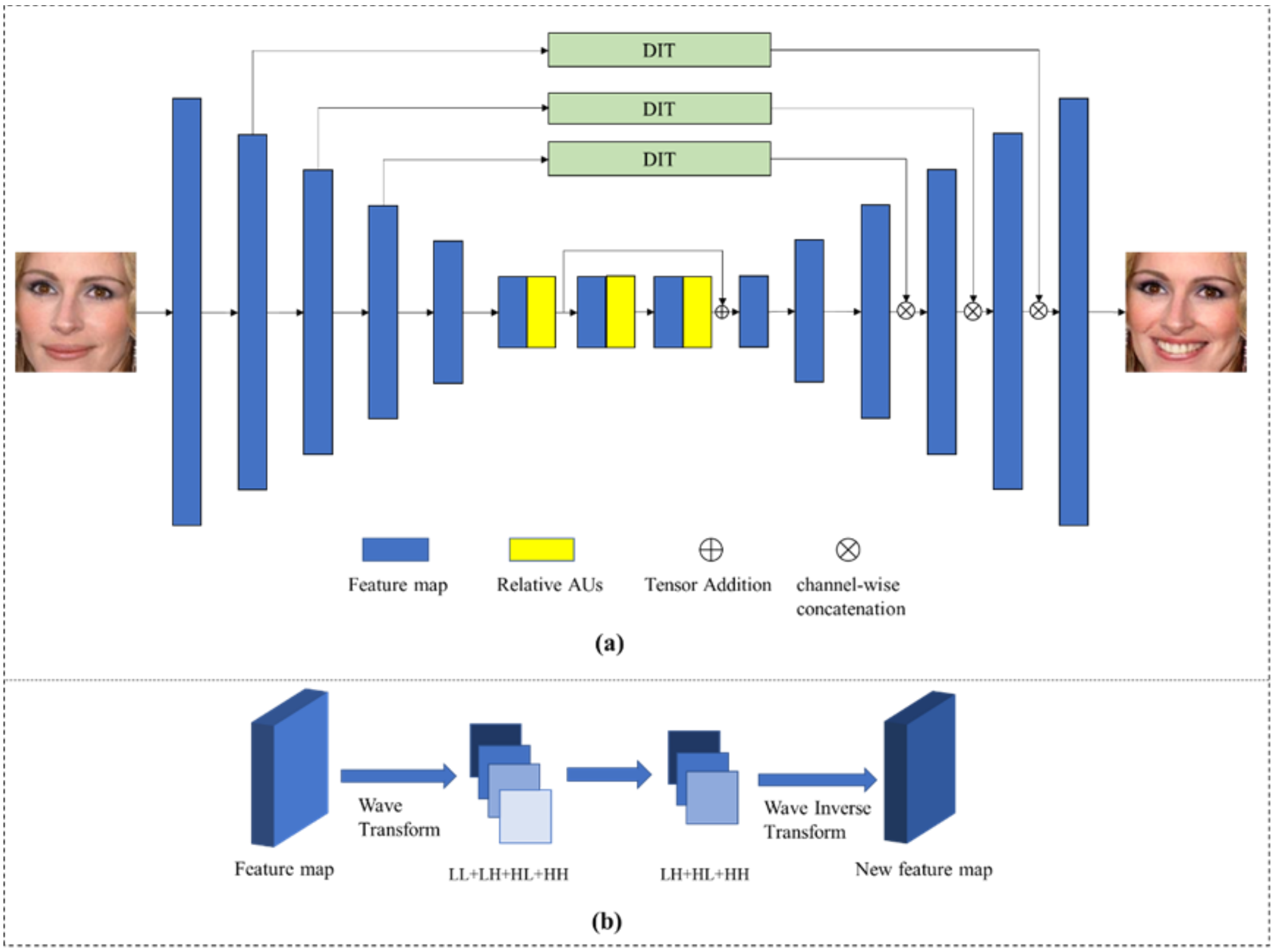

**Fig. 4** (a) Structure of the generator. The generator consists of three parts: encoder, decoder, and DIT module. The encoder and decoder are connected using residual blocks, and relative AUs are added several times to improve the editing capability of the network. (b) DIT module. The detail information of the feature map from the encoder is extracted using wavelet transform and transmitted to the decoder.

Fig. 4 (a) shows the details of our generator. The generator has the backbone of the U-net architecture and contains an encoder and decoder as well as a DIT module. The encoder contains six convolution layers, five convolution layers of which are for down-sampling to project the input to a low-dimensional space. The decoder consists of six corresponding convolution layers, five convolution layers of which for up-sampling to reconstruct the output image from the low-dimensional space. The DIT module uses wavelet transform to extract high-frequency components of the middle three feature maps of the encoder and transmits them to the decoder through a skip connection. In addition, a residual network block is used to connect the encoder and decoder, and relative AUs are introduced several times to improve the network model's ability to edit expressions. The structure of the generator is detailed in Table 1.

**Table 1** The structure of the generator network (Fig. 4(a))

| Type | Kernel | Stride | Padding | Normalization | Activation | Input Size | Output Size |
|---|---|---|---|---|---|---|---|
| | | | | Encoder | | | |
| Convolution | 7 | 1 | 3 | IN | LeakyReLU | 3×128×128 | 64×128×128 |
| Downsample | 4 | 2 | 1 | IN | LeakyReLU | 64×128×128 | 128×64×64 |
| Downsample | 4 | 2 | 1 | IN | LeakyReLU | 128×64×64 | 256×32×32 |
| Downsample | 4 | 2 | 1 | IN | LeakyReLU | 256×32×32 | 512×16×16 |
| Downsample | 4 | 2 | 1 | IN | LeakyReLU | 512×16×16 | 512×8×8 |
| Downsample | 4 | 2 | 1 | IN | LeakyReLU | 512×8×8 | 512×4×4 |
| | | | | Multiple AU Fusion Module | | | |
| Convolution | 3 | 1 | 1 | IN | ReLU | (512+17) ×4×4 | 512×4×4 |
| Convolution | 3 | 1 | 1 | IN | — | (512+17) ×4×4 | 512×4×4 |
| Convolution | 3 | 1 | 1 | IN | ReLU | 2 × (512+17) ×4×4 | 512×4×4 |
| | | | | Decoder | | | |
| Upsample | 4 | 2 | 1 | IN | ReLU | 512×4×4 | 512×8×8 |
| Upsample | 4 | 2 | 1 | IN | ReLU | 512×8×8 | 512×16×16 |
| Upsample | 4 | 2 | 1 | IN | ReLU | 2×512×16×16 | 256×32×32 |
| Upsample | 4 | 2 | 1 | IN | ReLU | 2×256×32×32 | 128×64×64 |
| Upsample | 4 | 2 | 1 | IN | ReLU | 2×128×64×64 | 64×128×128 |
| Convolution | 7 | 1 | 3 | — | Tanh | 64×128×128 | 3×128×128 |

We have two discriminators, denoted as $D_I$ and $D_H$ respectively. The discriminator $D_I$ contains a fully convolutional sub-network consisting of six convolutional layers with the kernel size of 4 and stride of 2. Then the multi-head mechanism is applied here with two branches. One branch adds a convolutional layer with kernel size 3, padding 1 and stride 1 to determine whether the image is fake or real, and the other branch adds an auxiliary regression head to predict the AUs label of the image. The discriminator $D_H$ is a high-frequency component discriminator, and the input is the high frequency component signal obtained from the wavelet transform of the image. The high-frequency component discriminator constrains the generator by determining whether the detail information distribution of the generated image aligns with the distribution of the original image, so that the generator could retain more high-frequency detail information of personal identity. The structure of the high-frequency component discriminator $D_H$ is shown in Table 2.

**Table 2** The structure of the high-frequency component discriminator $D_H$

| Type | Kernel | Stride | Padding | Normalization | Activation | Input Size | Output Size |
|---|---|---|---|---|---|---|---|
| Downsample | 4 | 2 | 1 | IN | LeakyReLU | 3×128×128 | 64×64×64 |
| Downsample | 4 | 2 | 1 | IN | LeakyReLU | 64×64×64 | 128×32×32 |
| Downsample | 4 | 2 | 1 | IN | LeakyReLU | 128×32×32 | 256×16×16 |
| Downsample | 4 | 2 | 1 | IN | LeakyReLU | 256×16×16 | 512×8×8 |
| Downsample | 4 | 2 | 1 | IN | LeakyReLU | 512×8×8 | 1024×4×4 |
| Downsample | 4 | 2 | 1 | IN | LeakyReLU | 1024×4×4 | 2048×2×2 |
| Convolution | 3 | 1 | 1 | - | - | 2048×2×2 | 1×2×2 |

## 3.3 Detail Information Transmission Module

The technique for extracting high frequency components of images is well established in the field of computer vision. The wavelet transform can decompose the image signal into different frequencies through low-pass and high-pass filters, where the high-pass filter allows only high-frequency information to pass, and the low-pass filter allows low-frequency information to pass. The two-dimensional discrete wavelet transform decomposes the original image into four frequency domain signals (LL, LH, HL, HH) by performing low-pass and high-pass filtering in horizontal and vertical directions, respectively. The LL signal can be taken as an approximation of the original image and consists of low-frequency information. The LH signal contains the detail information in the horizontal direction, the HL signal has the detail information in the vertical direction, and the HH signal is the detail information in the diagonal direction. The wavelet inverse transform can accurately reconstruct the original image using the signals in the four frequency domains obtained after the above wavelet transform decomposition, as shown in Fig. 4(b).

Following the idea of two-dimensional discrete wavelet transform, we used Haar wavelet transform [30], a classical method in wavelet transforms, as WEM-GAN to extract details. In Haar wavelet transform, low-pass filter $f_L = \left\{\frac{1}{\sqrt{2}}, \frac{1}{\sqrt{2}}\right\}$, and high-pass filter $f_H = \left\{-\frac{1}{\sqrt{2}}, \frac{1}{\sqrt{2}}\right\}$. Fig. 5 shows the effect of image decomposition using Haar wavelet transform. From Fig. 5, we can see that the high frequency details of the image are concentrated in the LH, HL, and HH components.

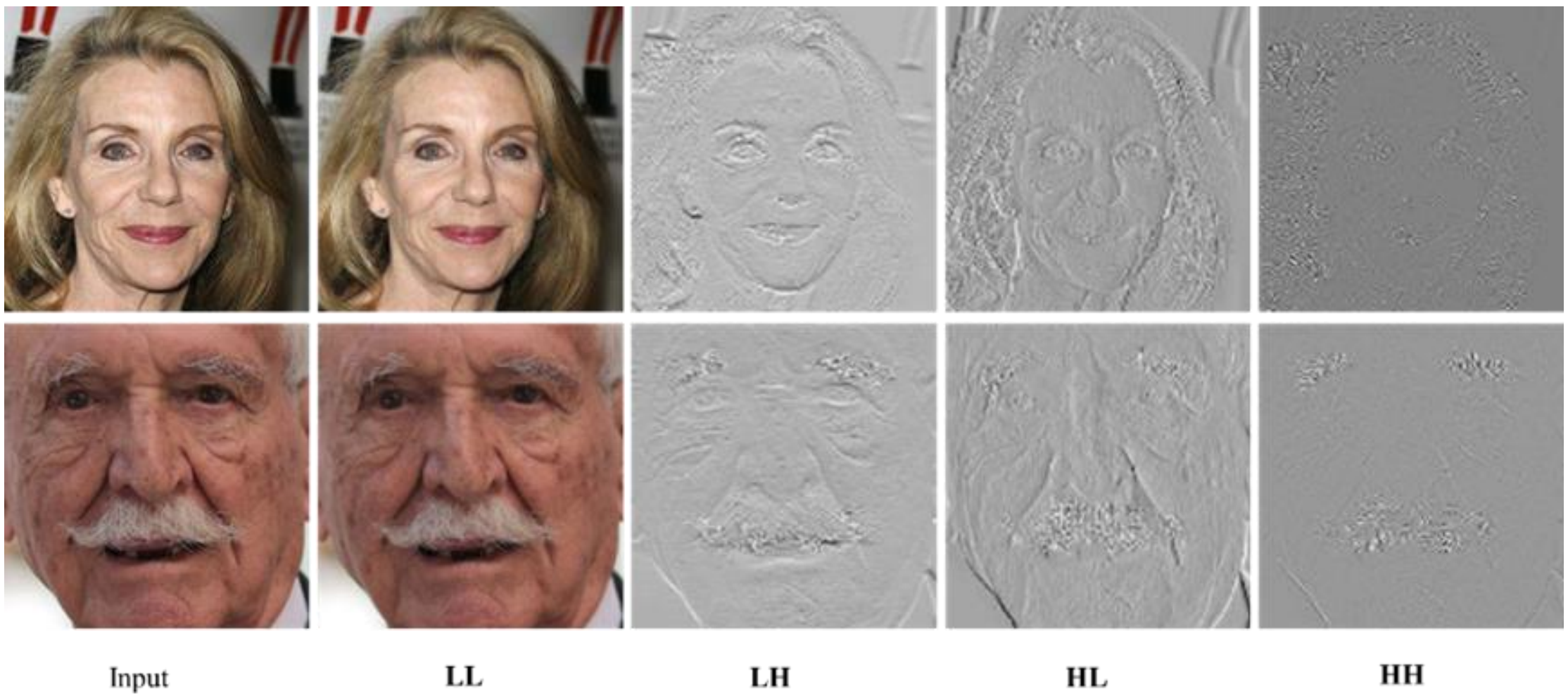

**Fig. 5** Haar wavelet transform visual effects

In WEM-GAN, the purpose of the DIT module is to extract the high-frequency components in the feature maps from encoder stages, and use them in the decoder to avoid the loss of detail information due to convolution and deconvolution layers. So, we perform Haar transform on the feature maps of different resolutions to four frequency domains, then we exclude the LL component signals and transmit the LH, HL, and HH signals to the decoder. At last, the high-frequency component signals are reconstructed into new feature maps containing only the high-frequency components using Haar inverse transform. The new feature map is channel-fused with the feature map from the decoder to provide more detail information for the decoder to reconstruct the image.

## 3.4 Loss Functions

Fig. 3 illustrates our training process. Given an input image x with its expression label $u_x$, and the target expression label $u_y$. We denote the relative AUs of the input $u_{rel} = u_y - u_x$, the generated image with the target expression as $x' = G(x, u_{rel})$, the generated self-reconstructed image as $x_{self} = G(x, 0)$, and the generated cycle-reconstructed image as $\hat{x} = G(x', -u_{rel})$. In addition, we denote the process of obtaining high frequency components from an image using wavelet transform as W.

The high frequency detail information of the input image and the output image can be denoted as $x_h = W(x)$ and $x'_h = W(x')$. The loss functions that we use in the model are described separately.

### 3.4.1 Adversarial Loss

To generate more realistic images from GANs, we extend the improved adversarial loss functions proposed by WGAN-GP [18]. Our adversarial loss can be written as:

$$\mathcal{L}_{D_{Iadv}} = -E_x[D_{Iadv}(x)] + E_{x'}[D_{Iadv}(x')] + \lambda_{gp} E_{\tilde{x}}[(\|\nabla_{\tilde{x}} D_{Iadv}(\tilde{x})\|_2 - 1)^2] \quad (1)$$

$$\mathcal{L}_{G_{adv}} = -E_{x,u_{rel}}\left[D_{Iadv}\left(G(x, u_{rel})\right)\right] \quad (2)$$

where $\lambda_{gp}$ is a penalty coefficient, and $\tilde{x}$ is a random interpolation between input x and the generated image $x'$.

### 3.4.2 High-Frequency Domain Adversarial Loss

In order to encourage the network to retain more detail information while generating images and ensure that the detail information distribution of the generated images is similar to that of the original images, we also use the generative adversarial loss as our high-frequency component loss function.

$$\mathcal{L}_{D_H} = -E_x[D_H(x_h)] + E_{x'}[D_H(x'_h)] + \lambda_{gp} E_{\widetilde{x_h}}\left[\left(\left\|\nabla_{\widetilde{x_h}} D_H(\widetilde{x_h})\right\|_2 - 1\right)^2\right] \quad (3)$$

$$\mathcal{L}_{G_H} = -E_{x,u_{rel}}\left[D_H\left(W\left(G(x, u_{rel})\right)\right)\right] \quad (4)$$

### 3.4.3 Expression Label Loss

In order to generate images with target expression, we follow the idea of GANimation and Ling et al. and add an auxiliary regressor $D_{Icond}$ to the last layer of the discriminator $D_I$ to detect the AUs labels of the images. We use the following loss functions to ensure the successful transformation of facial expressions.

$$\mathcal{L}_{D_{Icond}} = E_{x,u_x}[\|D_{Icond}(x) - u_x\|_2^2] \quad (5)$$

$$\mathcal{L}_{G_{cond}} = E_{x,u_{rel},u_y}\left[\left\|D_{Icond}\left(G(x, u_{rel})\right) - u_y\right\|_2^2\right] \quad (6)$$

### 3.4.4 Identity Information Loss

In order to preserve the identity information after expression editing, i.e., the faces in the input and output images can be recognized as the same person. We use the input image, the self-reconstructed image, and the cycle-reconstructed image to form the following loss to ensure the basic identity information after editing.

$$\mathcal{L}_{rec} = E_x\left[\left\|x - x_{self}\right\|_1\right] + E_{x,u_{rel}}[\|x - \hat{x}\|_1] \quad (7)$$

### 3.4.5 Objective Function

Combining all the loss functions mentioned above, the loss functions are summed up in a certain proportion to form the final loss function of our model, which is shown as follows:

$$\mathcal{L}_D = \mathcal{L}_{D_{Iadv}} + \mathcal{L}_{D_H} + \lambda_1 \mathcal{L}_{D_{Icond}} \quad (8)$$

$$\mathcal{L}_G = \mathcal{L}_{G_{adv}} + \mathcal{L}_{G_H} + \lambda_2 \mathcal{L}_{G_{cond}} + \lambda_3 \mathcal{L}_{rec} \quad (9)$$

Where $\lambda_1$, $\lambda_2$ and $\lambda_3$ are the hyperparameters to indicate the contribution of each loss to the final loss function.

# 4 Experiments

## 4.1 Implementation Details

**Dataset and Preprocessing.** The face image dataset we chose is AffectNet [31], which has more than 1 million images and 11 different categories such as Neutral, Happy, Sad, Surprise, Fear, Disgust, Anger, Contempt, None, Uncertain, Non-face. We randomly select a certain number of images from each of the first 10 categories to form our training set of 200,000 images and our test set of 2000 images to test our model. To obtain the expression label information of each image, we use the existing open-source tool OpenFace2.0 [32] to extract the AUs labels. In addition, all images are center-cropped and resized to 128*128.

To validate the robustness of our model, we also selected the FFHQ [33], RAF-DB [34], and CelebA [35] datasets for generalization testing, with the images similarly center-cropped to 128*128.

**Baseline.** We benchmark our model with GANimation [11] and Ling [12] et al. For better description, we name the model of Ling et al. as FGGAN. We use their open-source GitHub code in our benchmarking for fairness, and use the identical training and test sets for all methods. However, Wang [13] and Bodur [29] et al. do not provide their codes, so we could not evaluate their models and have to skip their methods in benchmarking results.

**Experiment Settings.** To ensure a fair comparison with the baseline model [12], we adopted its hyperparameter settings, with $\lambda_1$, $\lambda_2$, and $\lambda_3$ values set to 150, 150, and 30, respectively. Besides, we train the model using the Adam [36] optimizer with an initial learning rate of $1*10^{-4}$. The optimizer parameters are set to beta1=0.5 and beta 2=0.999. The model has a batch size of 16 and is trained for a total of 50 epochs. The model learning rate decreases linearly starting from the 31th epoch. For every 4 iterations of training the discriminator, the generator undergoes training once. The whole model training process takes around 96 hours on a 32G Tesla V100S-PCIE GPU.

## 4.2 Experiment Metrics

In order to quantitatively evaluate the performance of our model, we use several commonly used evaluation metrics based on our test set images and the generated target images.

**Inception Score (IS).** This metric uses Inception Net-V3 to output a 1000-dimensional vector for each input image [37], with each dimension representing the probability that the image belongs to a specific category. Statistically, these probabilities are calculated as the KL divergence of the conditional and marginal distributions, which can be used to measure the clarity and diversity of the generated images. The larger IS means the result is better. Although there are certain limitations of this metric, in this paper, we calculate the IS scores of the generated images following the evaluation methods of GANimation [11] and FGGAN [12]. The formula is as follows:

$$IS(x) = exp\mathbb{E}_{x\sim P_g}[KL(p(y|x)\|p(y))] \tag{10}$$

Here, $x$ represents the generated image, $P_g$ represents the distribution of the generated images, and $\mathbb{E}$ denotes the expectation over the samples. y represents the Inception classification result of the generated image, $p(y|x)$ denotes the output distribution of Inception when the generated image $x$ is input, and $p(y)$ represents the average distribution of the categories output by Inception for images generated by the generator $G$. exp denotes the exponential function.

**Frechet Inception Distance (FID).** This metric is an improved version of IS score [38]. It uses Inception Net-V3 network to extract the features of the image. In this paper, we use 2048-dimensional features from the last pooling layer and calculate the distance between the generated image features and the real image features. FID describes the distance between the two distributions. The smaller FID score indicates that the generated image is closer to the real image distribution, and the better the quality of the generated image is.

$$FID(x,g) = \|\mu_x - \mu_g\|_2^2 + Tr\left(\sum_x + \sum_g - 2\sqrt{\sum_x \sum_g}\right) \tag{11}$$

Here, $u_x$ and $\sum_x$ represent the mean and covariance matrix of the set of 2048-dimensional feature vectors detected from real images using the Inception Net-V3 network. Similarly, $u_g$ and $\sum_g$ represent the mean and covariance matrix of the feature vectors from the generated images. $Tr$ denotes the trace of a matrix.

**Average Content Distance (ACD).** ACD measures the L2 distance between the latent embeddings of the input and generated images [39]. We follow the approach used by GANimation [11] and FGGAN [12] to extract the facial embeddings of the input image and the generate image and calculate their distances using the well-known facial recognition network[1]. Lower values indicate better identity similarity between the images before and after editing.

$$ACD(x,g) = \|f_x - f_g\|_2 \tag{12}$$

Here, $x\ and\ g$ represent the input image and the generated image, respectively. $f_x$ and $f_g$ denote the facial encodings extracted using a facial recognition network from the input and generated images, respectively.

**Structure Similarity Index Measure (SSIM).** SSIM [40] measures the structure

[1] https://github.com/ageitgey/face_recognition

similarity between two images by comparing information such as brightness, contrast and structure between them, based on the assumption that the human eye can extract structural information from images. It can provide an evaluation of the image quality that is consistent with human perception. In this paper, we evaluate the ability of the model to preserve identity information by comparing the SSIM values between the original and self- reconstructed images.

$$SSIM(x,y) = \frac{(2\mu_x\mu_y + C_1)(2\sigma_{xy} + C_2)}{(\mu_x^2 + \mu_y^2 + C_1)(\sigma_x^2 + \sigma_y^2 + C_2)} \tag{13}$$

Here, $x$ and $y$ are two images, $\mu_x$ and $\mu_y$ represent the mean pixel values of the images, $\sigma_x$ and $\sigma_y$ denote the standard deviation of the pixel values, and $\sigma_{xy}$ represents the covariance between $x$ and $y$. $C_1$ and $C_2$ are constants.

**Expression Distance (ED).** Because the editing of expressions is continuous, we cannot simply classify the edited expressions into discrete categories such as Happy, Sad, etc. to calculate the classification accuracy of expression editing models. To determine whether the generated images have the target expressions, we also use OpenFace2.0 to extract the AUs of the generated images and calculate the $l_2$-distance between them and the target AUs as the expression distance to approximate the model's ability to edit expressions.

$$ED = \mathbb{E}_{x\sim P_g}\left[\left\|\bar{u}_x - u_y\right\|_2\right] \tag{14}$$

Here, $x$ represents the generated image, $P_g$ denotes the distribution of the generated images, and $\mathbb{E}$ denotes the expectation over the samples. $\bar{u}_x$ is the AU label vector detected from the generated images using OpenFace 2.0, and $u_y$ is the target AU label vector.

**Face Verification Score (FVS).** We use Face++[1] to computes the similarity between input and synthesized images, it will return a value between 0 and 100 to indicate the likeness between two faces.

## 4.3 Qualitative Evaluation

We start by comparing the AU editing performance with GANimation and FGGAN. Fig. 6 shows the editing results for the two classical AU labels AU5 (Upper Lid Raiser) and AU45 (Blink), and Fig. 7 shows the editing results (the AU combination of Happy expressions) for the combination of AU6 (Cheek Raiser) + AU12 (Lip Corner Puller).

[1] https://www.faceplusplus.com/

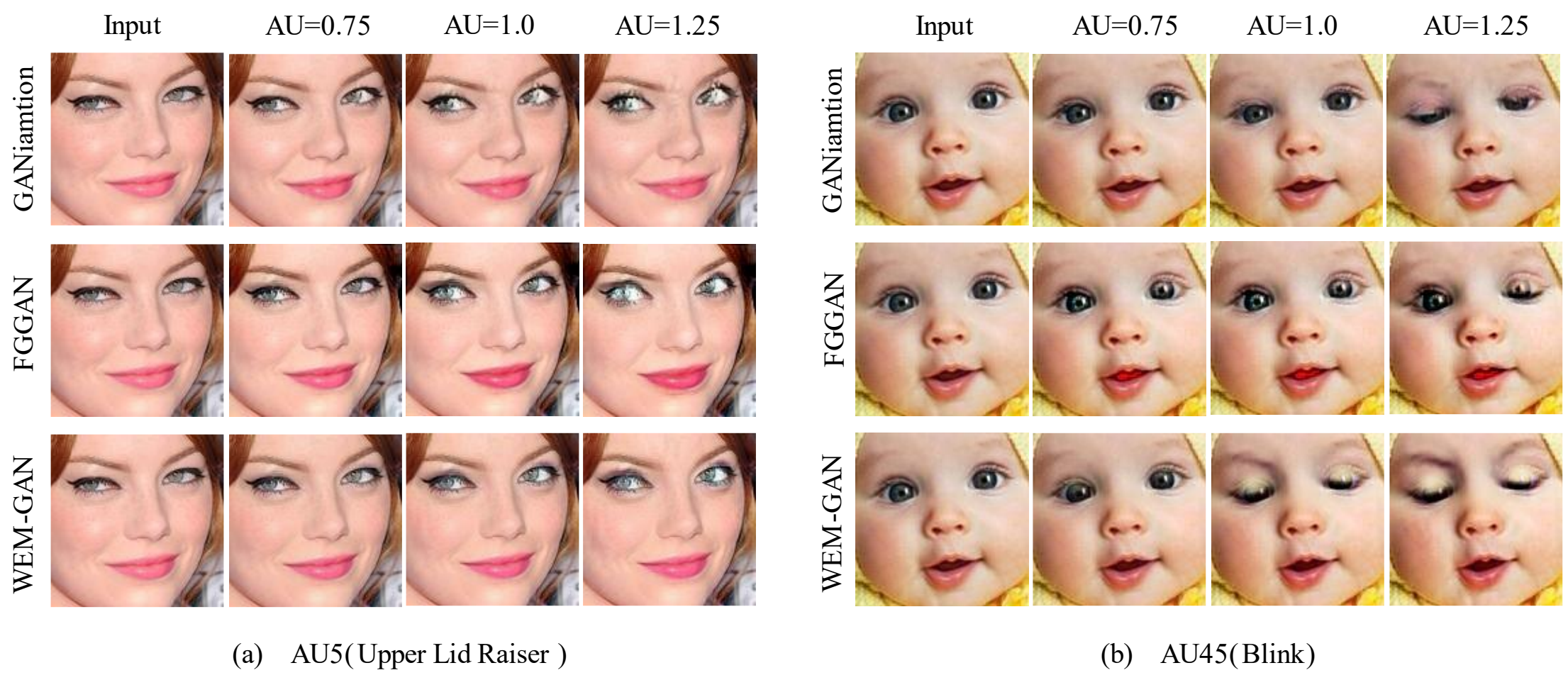

**Fig. 6** Comparison of expression editing results for single AU label. We manipulate the facial expression with only one AU label. The given values denote the intensity of relative AUs value used in our test.

From (a) of Fig. 6, we can discern that the output of GANimation changes less when we tune the AU5 value, and the generated image has artifacts in the eye region. FGGAN and our results are able to make corresponding changes in expression according to the changes in AU, and have similar performance in terms of capturing AU. However, our results are clearer in the eye region, and almost no changes occurred at other irrelevant face regions. From (b) of Fig. 6, we can understand that GANimation and FGGAN do not generate the expression successfully when the value of AU45 changes, and the generated images produce unnecessary artifacts in other irrelevant regions (e.g., FGGAN mouth region), while our generated images not only succeed in closing the eyes according to the AU45 labels, but also are realistic, clear, and natural.

From Fig. 7, we can conclude that when multiple AUs are edited simultaneously, the images generated by GANimation are blurrier and have more artifacts. FGGAN generates images that are still clear when the magnitude of change is not large, but when the value changes drastically such as in the last column, it also generates artifacts in irrelevant areas. However, compared to the first two methods, our generated images can be modified continuously according to the change of AU value, and the intermediate results are all clear and realistic.

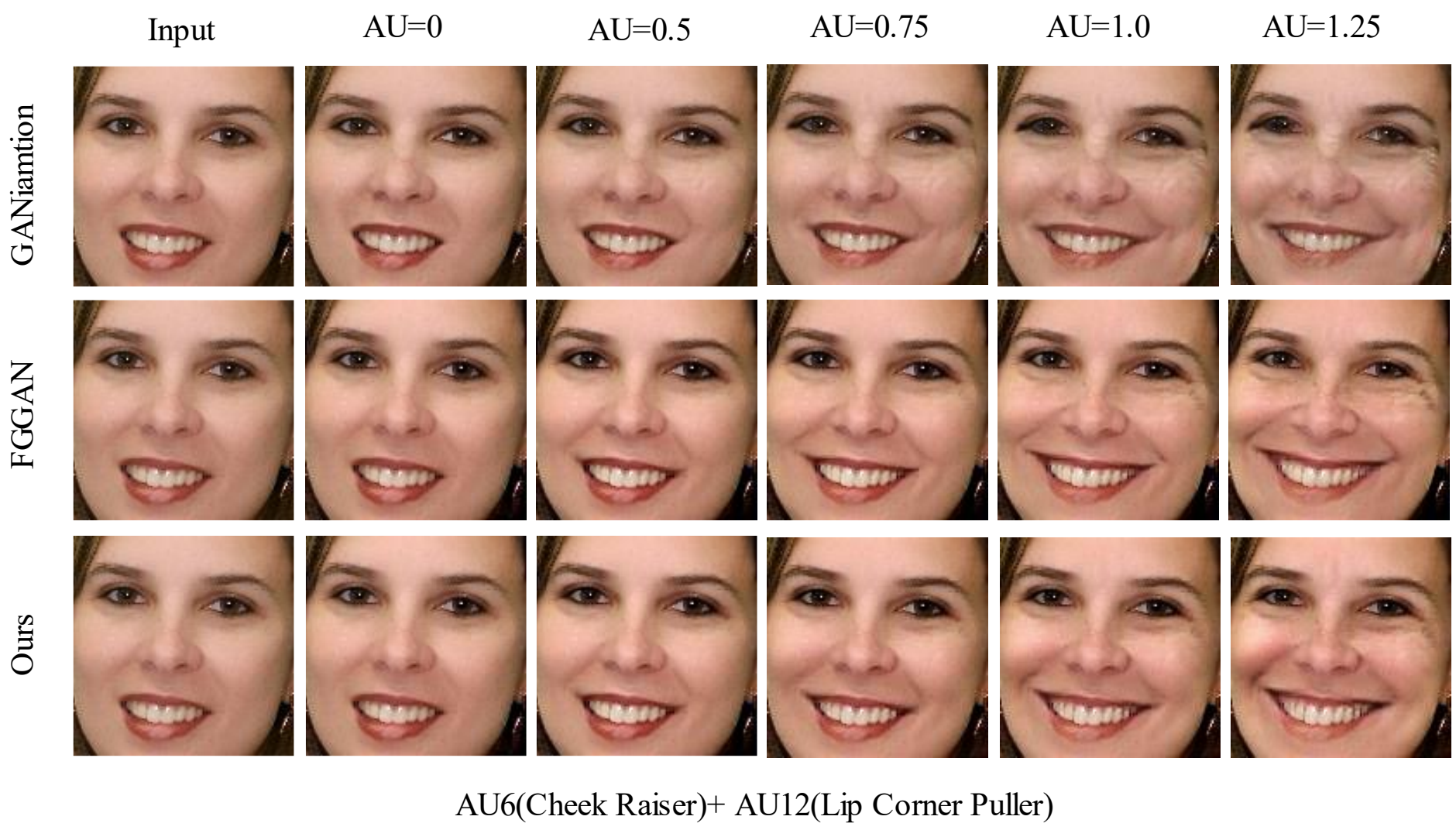

**Fig. 7** Comparison of expression editing results for combination AU labels. We manipulate the face with both AU6 and AU12. The given values upon images denote the intensity of relative AUs.

Additionally, we select one image as the target expression from each of the eight expression categories such as Neutral, Happy, Sad, Surprise, Fear, Disgust, Anger, and Contempt. Expression conversion is performed on the test set, and the results are shown in Fig. 8 FGGAN produces fewer artifacts than GANimation but does not preserve the face details of the original image well in areas not related to expression transformation (left cheek in the third row on the left, and lip color in the second row on the right). Compared with the two methods, our model not only successfully performs expression transformation but also does a better job in keeping identity information, producing less artifacts, and generating more natural and realistic outputs.

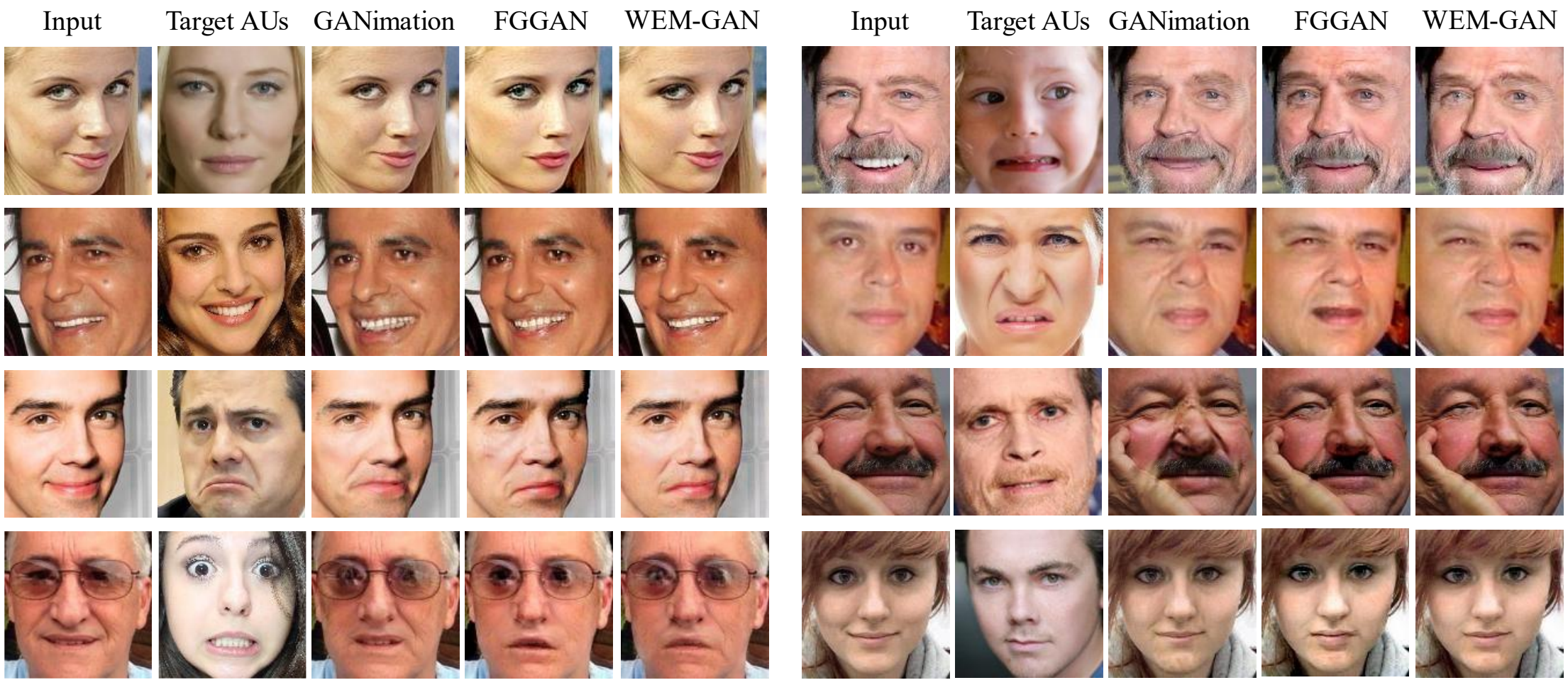

**Fig. 8** Generated images with target expression comparison

Lastly, to verify the model's ability of continuous expression editing, AU interpolation between two target expressions is performed, and the editing results are shown in Fig. 9, where the expressions of the input images from left to right are gradually transformed from target expression 1 to target expression 2. From Fig. 9, it can be observed that WEM-GAN can continuously modify expressions according to AUs values, which also validates the superiority of using AUs as the expression label.

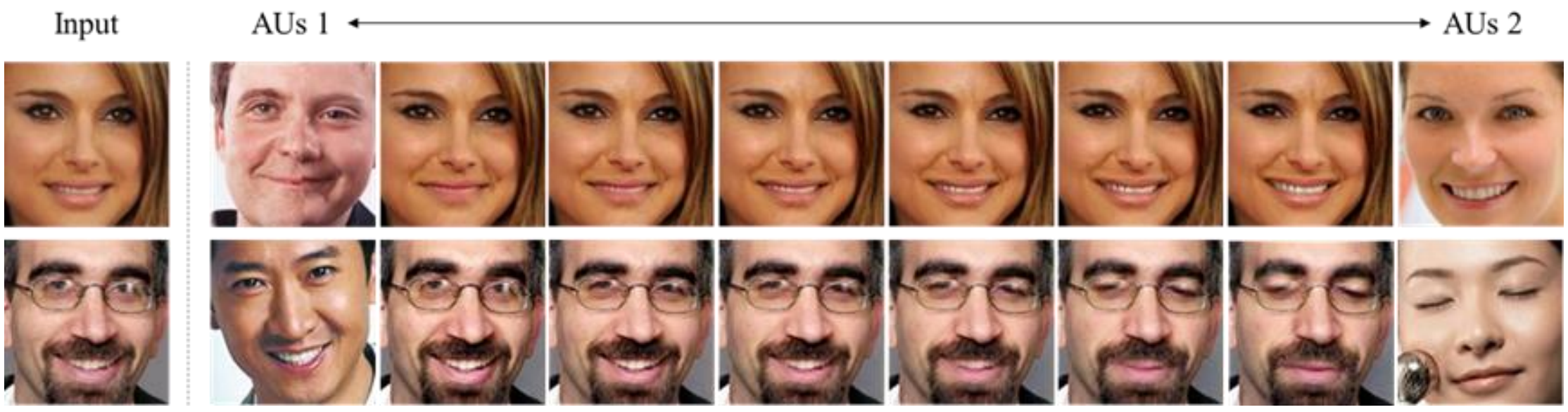

**Fig. 9** Interpolation results of continuous expression editing

## 4.4 Quantitative Results

In this section, we will use the metrics proposed in Section 4.2 to benchmark three models. We randomly select 2000 images as our test set. To increase the fairness of our benchmarking, each image is randomly selected with another image in the test set as the target expression for transformation, and we ensure that the combination of input image and target expression image are identical in all three models when testing.

We start by calculating the ACD and ED between the input images and the images after expression editing, and the results are shown in Table 3. Our model could generate images with more detail information by focusing on keeping more details and gets the lowest ACD scores compared to GANimation and FGGAN. Combining the residual network with multiple relative AUs also enhances the network ability to transform facial expressions and obtains competitive results in terms of ED scores, which suggests that our model can achieve a more fine-grained level of facial expression editing.

**Table 3** ACD and ED result

| Method | ACD↓ | ED↓ |
|---|---|---|
| GANimation[11] | 0.3601 | 0.2113 |
| FGGAN[12] | 0.3549 | 0.2004 |
| Wang[13] | — | 0.245* |
| Bodur[29] | — | — |
| EGA-GAN[41] | 0.307* | 0.480* |
| EmoStyle[42] | — | — |
| WEM-GAN | **0.3197** | **0.1943** |

*Note: Because the author neither open-source the code nor the dataset, it is hard to train and test the model with the same settings. So, we directly borrow the best experimental results from their original paper. The following *-marked data describes the same situation. However, since the dataset used by the author in their experiment are different from WEM-GAN, the results are only for reference.

Additionally, to further understand the ability of our model to retain identity information, we set the relative AUs to 0 and generate a self-reconstructed image related to the input image which can be viewed as the ground truth information of the self-reconstructed image. The ability of the model to preserve identity information before and after expression editing is measured by calculating the distance between the input image and the self-reconstructed image. In our work, we use L1, PSNR (Peak Signal-to-Noise Ratio) and SSIM distance as our metrics for evaluation. The results shown in Table 4 suggest that, compared with our model, GANimation and FGGAN have achieved better results in terms of all three metrics. However, Wang et al did not provide any open-sourced codes and datasets, the experiment results are not benchmarked by training and testing the model with the same dataset. Thus, it cannot be clearly explained that our model is inferior to that of Wang et al**.**

**Table 4** Reconstruction comparison

| Method | L1↓ | SSIM↑ | PSNR↑ |
|---|---|---|---|
| GANimation[11] | 0.0174 | 0.9485 | 30.80 |
| FGGAN[12] | 0.0145 | 0.9768 | 33.87 |
| Wang[13] | **0.008*** | **0.9953*** | — |
| Bodur[29] | — | 0.84* | 35.29* |
| EGA-GAN[41] | — | — | — |
| EmoStyle[42] | — | — | — |
| WEM-GAN | 0.0129 | 0.9829 | **35.60** |

Finally, we use the IS score and FID score to measure the quality of the images generated by our model, and the results are shown in Table 5. Our model pays more attention to the image detail information and generates images with fewer artifacts, better clarity, and better image quality, so it also achieves the best for both IS and FID metrics.

**Table 5** IS and FID distance

| Method | IS↑ | FID↓ |
|---|---|---|
| Real Images | 2.2797 | — |
| GANimation[11] | 2.1810 | 7.28 |
| FGGAN[12] | 2.1818 | 7.63 |
| Wang[13] | — | 8.51* |
| Bodur[29] | — | 13.35* |
| EGA-GAN[41] | — | 14.51* |
| EmoStyle[42] | — | 7.86* |
| WEM-GAN | **2.2063** | **6.73** |

## 4.5 Ablation Study

In this section, we explore the importance of each component of our proposed method. For the convenience of description, the residual network block in the middle of the encoder and decoder, which fuses the relative AUs multiple times, is named the multiple AU fusion module and denoted by Mul_AU, and the high-frequency component discriminator is denoted by $D_h$. We examine these proposed modules using FID, ACD, and ED metrics. From the initial U-Net backbone generator, the three modules are gradually added with relative AUs as the conditional vector, and each model is trained and tested with the same data set. The experimental results are shown in Table 6.

**Table 6** Ablation study results

| Models | FID↓ | ACD↓ | ED↓ |
|---|---|---|---|
| GANimation[11] | 7.28 | 0.3601 | 0.2113 |
| FGGAN[12] | 7.63 | 0.3549 | 0.2004 |
| WEM-GAN w/o Mul_AU+DIT+$D_h$ | 22.82 | 0.3446 | 0.2038 |
| WEM-GAN w/o DIT+$D_h$ | 21.44 | 0.4198 | 0.1987 |
| WEM-GAN w/o $D_h$ | 7.09 | 0.3209 | 0.1964 |
| WEM-GAN | **6.73** | **0.3197** | **0.1943** |

As it can be seen in Table 6, with the adding of the multiple AU fusion module, the ability of the model to retain detail information decreases as the number of network levels in the generator increases, leading to a rise in the ACD metric. With the adding of wavelet transform techniques in the generator, there is a significant decrease in the ACD and FID metrics. The high-frequency component discriminator further enhances the generator's ability to keep detail information and generate images with more details, and it achieves better performance in ED when combined with the proposed multiple AU fusion module. From the experimental results, it can be concluded that our proposed detail information transmission module provides significant improvements in retaining the detail information of the images.

We also explored the levels at which skip connections are introduced in the generator. During the encoding process, high-dimensional feature maps contain richer identity information, while low-dimensional ones contain less. Therefore, we selected the middle three layers as connection layers to simplify the network while achieving good results. Ablation experiments were conducted using the three metrics mentioned above to evaluate the network models with skip connections at different layers. The results, shown in Table 7, indicate that adding more connection layers does not significantly improve the results and makes the network structure more complex. Hence, considering these factors, using the middle three layers for connections already achieves good results.

**Table 7** Comparison of ablation study results for skip connection layers

| Method | FID ↓ | ACD ↓ | ED ↓ |
|---|---|---|---|
| WEM-GAN w/o DIT | 21.44 | 0.4198 | 0.1987 |
| WEM-GAN w DIT, 5 layers | 6.83 | **0.3112** | 0.1949 |
| WEM-GAN w DIT, 4 layers | 6.77 | 0.3133 | 0.1985 |
| WEM-GAN w DIT, 3 layers | **6.73** | 0.3197 | **0.1943** |

## 4.6 Generalization

We trained our model on the AffectNet dataset and tested it on other datasets, such as FFHQ [33], RAF-DB [34], and CelebA [35], to validate the generalization capability of our model. The results are shown in Fig. 10 and 11.

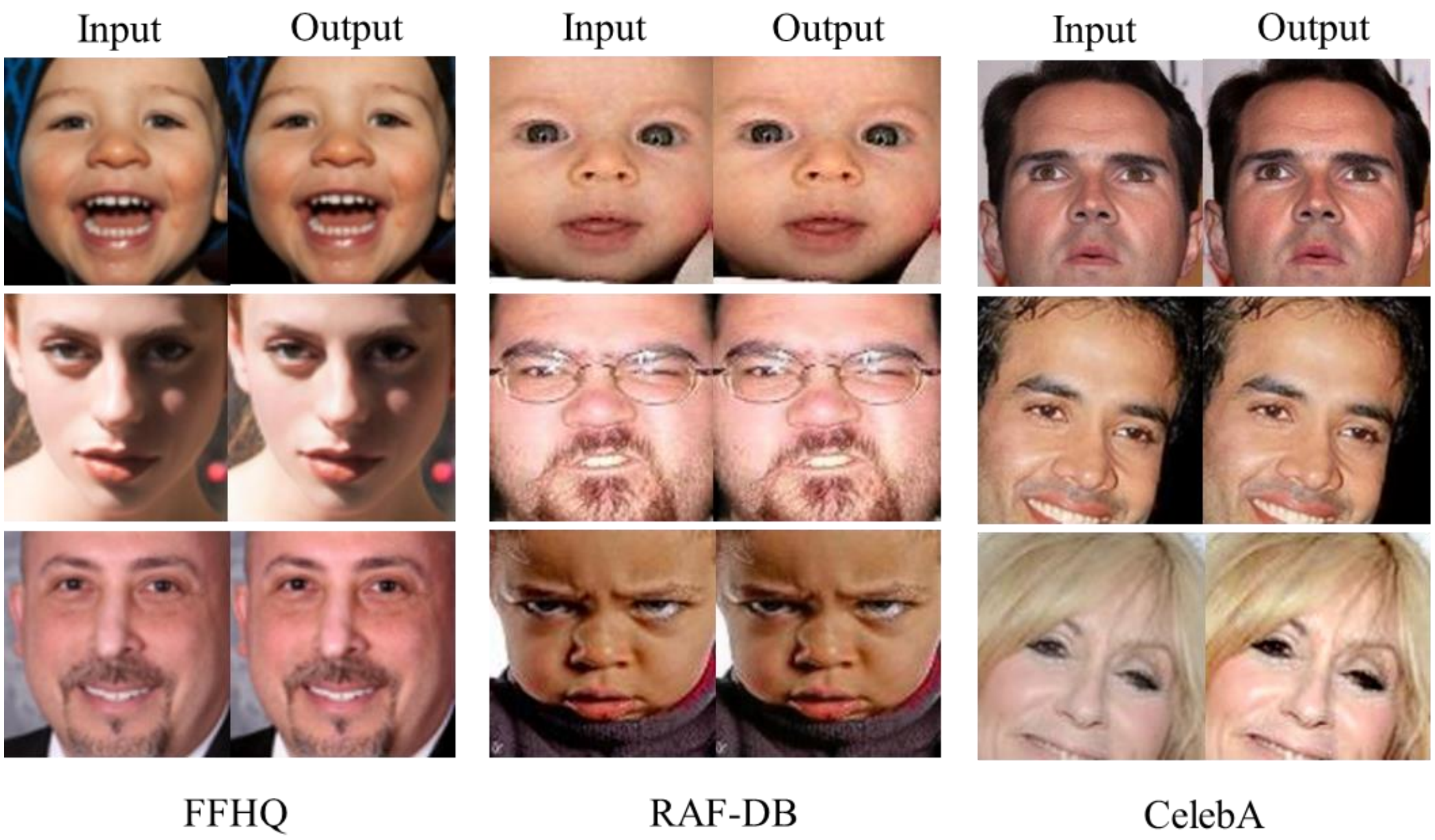

**Fig. 10** Reconstruction ability on other datasets

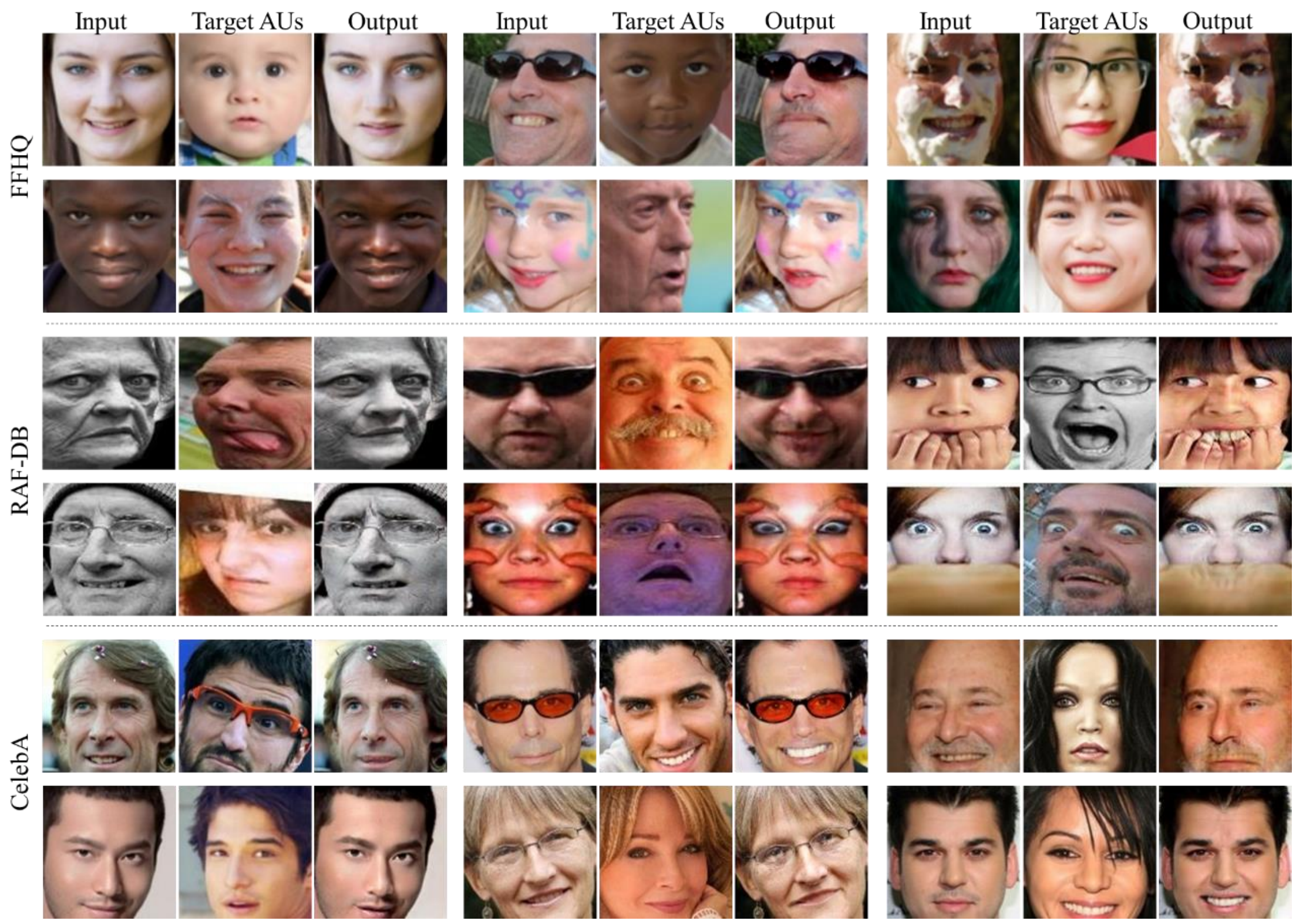

**Fig. 11** Generated images with target expression on other datasets

From Fig. 10 and 11, we can see that our method performs well on other datasets. Fig. 10 illustrates that WEM-GAN can realistically reconstruct the input images, while Fig. 11 demonstrates that WEM-GAN can achieve realistic expression editing effects even on some extreme cases.

Additionally, we calculated some qualitative metrics for WEM-GAN on these datasets. The reconstruction and editing results are shown in Tables 8 and 9, respectively.

**Table 8** Quantitative comparison of reconstruction performance on different datasets

| Dataset | L1 ↓ | SSIM ↑ | PSNR ↑ |
|---|---|---|---|
| FFHQ | 0.0253 | 0.9559 | 32.86 |
| RAF-DB | 0.0226 | **0.9584** | 33.81 |
| CelebA | 0.0225 | 0.9526 | 33.45 |
| AffectNet | **0.0129** | 0.9829 | **35.60** |

**Table 9** Quantitative comparison of editing results on different datasets

| Dataset | FID ↓ | ACD ↓ | ED ↓ | IS ↑ | FVS ↑ |
|---|---|---|---|---|---|
| FFHQ | 11.33 | 0.3470 | 0.1985 | **2.9136** | 91.84 |
| RAF-DB | 13.56 | **0.3051** | **0.1859** | 2.7661 | **93.16** |
| CelebA | 7.66 | 0.3359 | 0.1862 | 2.5063 | 92.67 |
| AffectNet | **6.73** | 0.3197 | 0.1943 | 2.2063 | 93.11 |

From Table 8 and 9, it can be seen from the quantitative metrics that even without fine-tuning on data from other datasets, WEM-GAN achieves excellent results. This demonstrates the robustness of our method, which can handle various complex situations present in different datasets and still achieve good results.

# 5 Conclusion and Future Work

In this study, we propose a new approach, WEM-GAN, to introduce wavelet transform technique for facial expression editing based on the U-Net backbone. In the generator, a wavelet transform-based detail information transmission module is implemented to address the information loss when the image is embedded in a low-dimensional latent space, and additionally add a high-frequency component discriminator to constrain the network model to further ensure the retention of image detail information. In addition, the residual network blocks with multiple fused relative AUs are used to connect the encoder and decoder to enhance the model's ability to edit facial expressions. The qualitative and quantitative experimental results show that our model not only performs well in preserving personal identity information, but also shows competitive performance in terms of facial expressions editing and the quality of the generated images.

Additionally, as a potential direction for future work, we plan to explore integrating temporal information into our model to achieve video-based facial expression manipulation. This aims to maintain the consistency and fidelity of expressions across consecutive video key frames, enhancing the adaptability and practicality of WEM-GAN in other real-world applications.

# Conflict of Interest

The authors declare that they have no conflict of interest.

# Data Availability

Data sharing is not applicable to this article as no datasets were generated during the current study.